%% file: root.tex
\documentclass[conference]{IEEEtran}
\IEEEoverridecommandlockouts
\usepackage{cite}
\usepackage{amsmath,amssymb,amsfonts}
\usepackage{algorithmic}
\usepackage{graphicx}
\usepackage{textcomp}
\usepackage{xcolor}
\usepackage{csquotes}
\usepackage{gensymb}
\usepackage{makecell}

\def\BibTeX{{\rm B\kern-.05em{\sc i\kern-.025em b}\kern-.08em
    T\kern-.1667em\lower.7ex\hbox{E}\kern-.125emX}}
\begin{document}

\title{SEDRo: A Simulated Environment for Developmental Robotics}

\author{\IEEEauthorblockN{Aishwarya Pothula, Md Ashaduzzaman Rubel Mondol, Sanath Narasimhan, Sm Mazharul Islam, Deokgun Park}

\IEEEauthorblockA{\textit{Computer Science and Engineering} \\
\textit{University of Texas at Arlington}\\
Arlington, Texas USA \\
\{aishwarya.pothula, mdashaduzzaman.mondol, sanath.narasimhan, sxi7321\}@mavs.uta.edu, deokgun.park@uta.edu}
}


\maketitle

\begin{abstract}
Even with impressive advances in application
specific models, we still lack knowledge about how to build a model that can learn in a human-like way and do multiple tasks.
To learn in a human-like way, we need to provide a diverse experience that is comparable to human's.
In this paper, we introduce our ongoing effort to build a simulated environment for developmental robotics (SEDRo). SEDRo provides diverse human  experiences ranging from those of a fetus to a 12th month old. A series of simulated tests based on developmental psychology will be used to evaluate the progress of a learning model.  
We anticipate SEDRo to lower the cost of entry and facilitate research in the developmental robotics community.  
\end{abstract}


\input{sections/introduction}
\input{sections/background}

\input{sections/plan}

\input{sections/description}
\input{sections/discussion}

\bibliography{references}
\bibliographystyle{IEEEtran}

\end{document}

%% file: sections/introduction.tex
\section{Introduction}

Imagine a robot that can work as a butler. It can handle many tasks and talk with other butler robots to do even more tasks. Alas, one cannot buy or build one today even with an unlimited budget. The reason a butler robot is not available is because we do not know how to program it. Current  approaches require huge data to teach a single skill~\cite{lake2017building}, and the data requirement grows exponentially with the number of tasks. While we have made remarkable progress in solving tasks with well-defined structures such as when explicit rewards or ground truth exist, we do not know how we can generalize this capability for a single task to multiple tasks.   Turing suggested~\cite{turing1950computing}: 

\begin{displayquote}
Instead of trying to produce a programme to simulate the adult mind, why not rather try to produce one which simulates the child's?
\end{displayquote}

\begin{figure}[b]
  \centering
  \resizebox{\columnwidth}{!}{\includegraphics{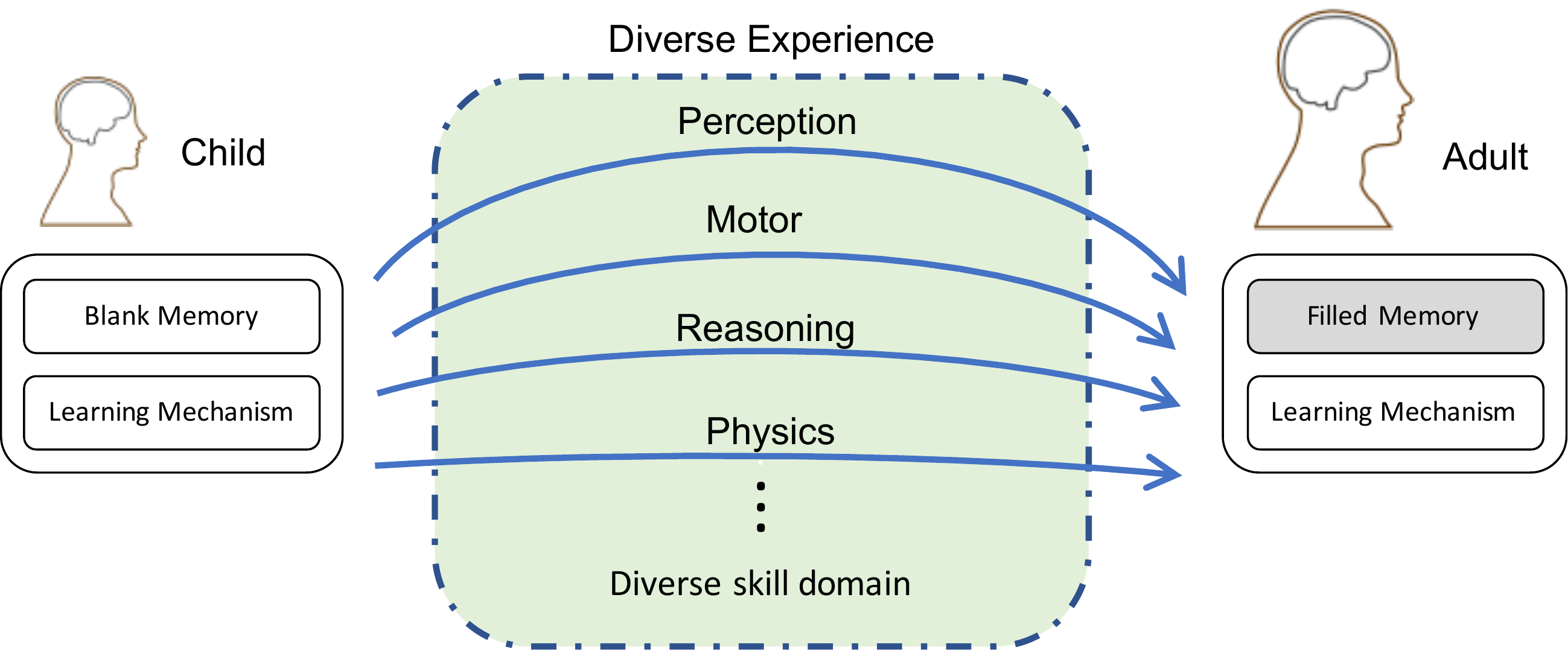}}
  \caption{\small A child begins with vast blank memory and a learning mechanism. The learning mechanism uses diverse experiences to fill memory. There can be many different skill domains, such as perception, motor, reasoning, and so on.  }
  \label{fig:example}
\end{figure}

Humans are born with a vast blank memory and a mechanism for filling it. Let us call this mechanism \textit{the learning mechanism} in this paper. With diverse experiences as input, the mechanism fills the contents of the memory as shown in Fig.~\ref{fig:example}. After a few years, we can do many things in multiple domains such as perception, motor, social, language, and physics. 
We claim that there were the following issues in previous approaches that made the search for the learning mechanism difficult and propose a new approach to mitigate those issues. 

\begin{itemize}
    \item  \textbf{Targeting a single skill rather than diverse skills (the main issue)} --  While a human child can learn to do many things simultaneously, we have focused largely on developing models that can do only a single task.  This approach had resulted in overfitted solutions that cannot be generalized to diverse tasks.
    \item \textbf{Use of refined and focused datasets rather than diverse and noisy datasets (the first common pattern)} -- Because the focus is to teach one skill, we tend to build a refined dataset or an environment that contains only task relevant information. This resulted in \textit{spoon-fed human-edited sensory data}~\cite{weng2001autonomous}. Compare this with how humans learn from unstructured data such as visual and auditory senses and find underlying structures and apply these structures to many domains\cite{gopnik1999scientist}. 
    
     \item \textbf{Relying on explicit rewards rather than on other mechanisms (the second common pattern)} -- While operant conditioning is a powerful mechanism~\cite{skinner1938behavior}, we tend to rely on explicit rewards to guide learning. 
    Designing a reward mechanism might be easy for a single task. However, it becomes exponentially difficult as the number of target tasks increase. 
    If we compare the language acquisition abilities of humans and robots, robots can learn to navigate according to the verbal instruction quickly~\cite{hermann2017grounded, chaplot2018gated, chen2019touchdown} but do not know how to generalize this to other tasks such as cooking. On the contrary, human infants cannot follow verbal instruction for a very long time. Unfortunately, you cannot give a treat to an 8-month infant for toilet training when he goes to a bathroom himself. But slowly around two years when they acquire language, they can do many tasks with it~\cite{gopnik1999scientist}.  One key difference is that while robots are trained using explicit rewards, it is not the case with infants.
    
\item \textbf{Too many necessary components rather than a sufficient set of the learning mechanism (the third common pattern)} -- Finally, we tend to find individual necessary mechanisms rather than suggesting a set of the sufficient mechanisms. The learning mechanism is a system of multiple components. Some might classify the components into two different categories: 1) innate or built-in mechanisms versus 2) universal principles that drive learning. Examples of innate mechanisms are reflexes, hippocampus, or limbic systems. Universal principles explain the driving force behind learning and can be usually written as succinct mathematical formulation such as intrinsic motivation~\cite{oudeyer2007intrinsic,schmidhuber2010formal}, Bayesian statistics~\cite{gopnik2004theory}, or the free energy principles~\cite{friston2010free}. As we can see, there are many candidate components, and we anticipate that the learning mechanism will be  a set of multiple components. 
    However, for a single application, a single or small subset of these components might do the job.   
    The problem is that we cannot linearly concatenate the solutions from multiple domains because they are not independent. 
    Therefore, a more critical but neglected question is what is a sufficient set of components for all problems humans can solve.

\end{itemize}

As a summary, we tend to build models for single tasks resulting in overfitted solutions that cannot be generalized to multiple tasks. 
In this perspective, we need a regularization. 
Regularization by sharing is an effective pattern as demonstrated in convolutional neural network (CNN) or recurrent neural network (RNN)~\cite{srivastava2014dropout}. 
We claim that we need to regularize by enforcing the use of the same learning mechanism to conduct multiple tasks as Allen Newell suggested in his unified theories of cognition~\cite{newell1994unified}.

Then why has the focus of past researches been on developing models for individual tasks? Imagine that a researcher has decided to build an agent that can perform many tasks like a human can. 
The first problem she encounters is that there is no simulated environment that can provide the diverse experiences required to acquire skills across multiple domains. 



To solve this problem, we introduce our ongoing effort to build a Simulated Environment for Developmental Robotics (SEDRo). SEDRo provides diverse experiences similar to that of human infants from the stage of a fetus to 12 months of age. SEDRo also simulates developmental psychology experiments to evaluate the progress of intelligence development in multiple domains.

\begin{figure}[tb]
  \centering
  \resizebox{0.95\columnwidth}{!}{\includegraphics{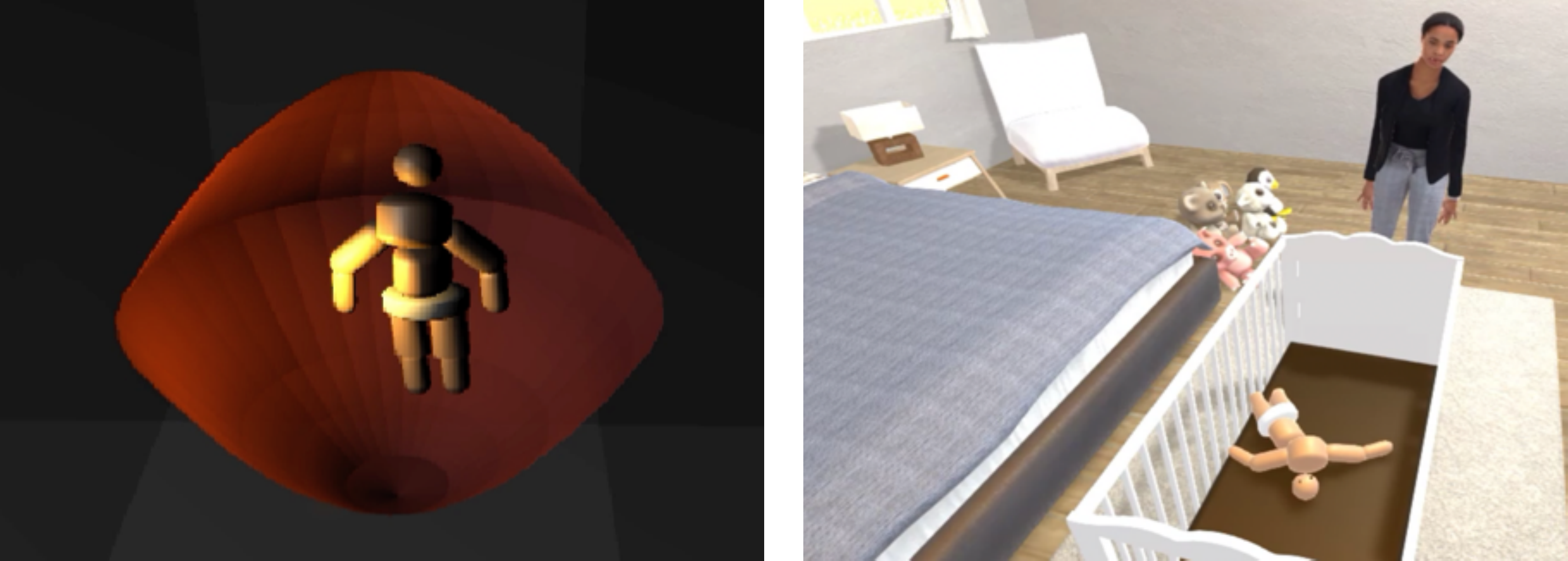}}
  \caption{\small Screenshots of SEDRo environment. The environment simulates a fetus in the womb(Left).  An infant (learning agent)  in the crib with a mother character (right).}
  \label{fig:environment}
\end{figure}

There are two generalizable lessons in our work. 
First, we point that the learning environment should provide experiences for the multiple tasks and provide a proof-of-concept example.  
Fig.~\ref{fig:environment} shows screenshots of SEDRo.
In our environment, the learning agent has to rely on interactions with other characters such as a mother character, who teaches language as a human mother does.
Other characters have to intelligently react to the random babbling of the baby in a diverse but reasonable way. 
Programming a mother character for all situations is intractable and it becomes increasingly challenging to provide an experience for open-ended learning when social learning is involved. 
In our paper, we address this issue by focusing on the earlier stage of development from the stage of a fetus to 12 months of age when a few words are acquired.  
It is more tractable as the conversations between the mother and the baby tends to be one-directional rather than interactive back-and-forth conversations. 

\begin{figure*}[htb]
  \centering
  \resizebox{0.91\textwidth}{!}{\includegraphics{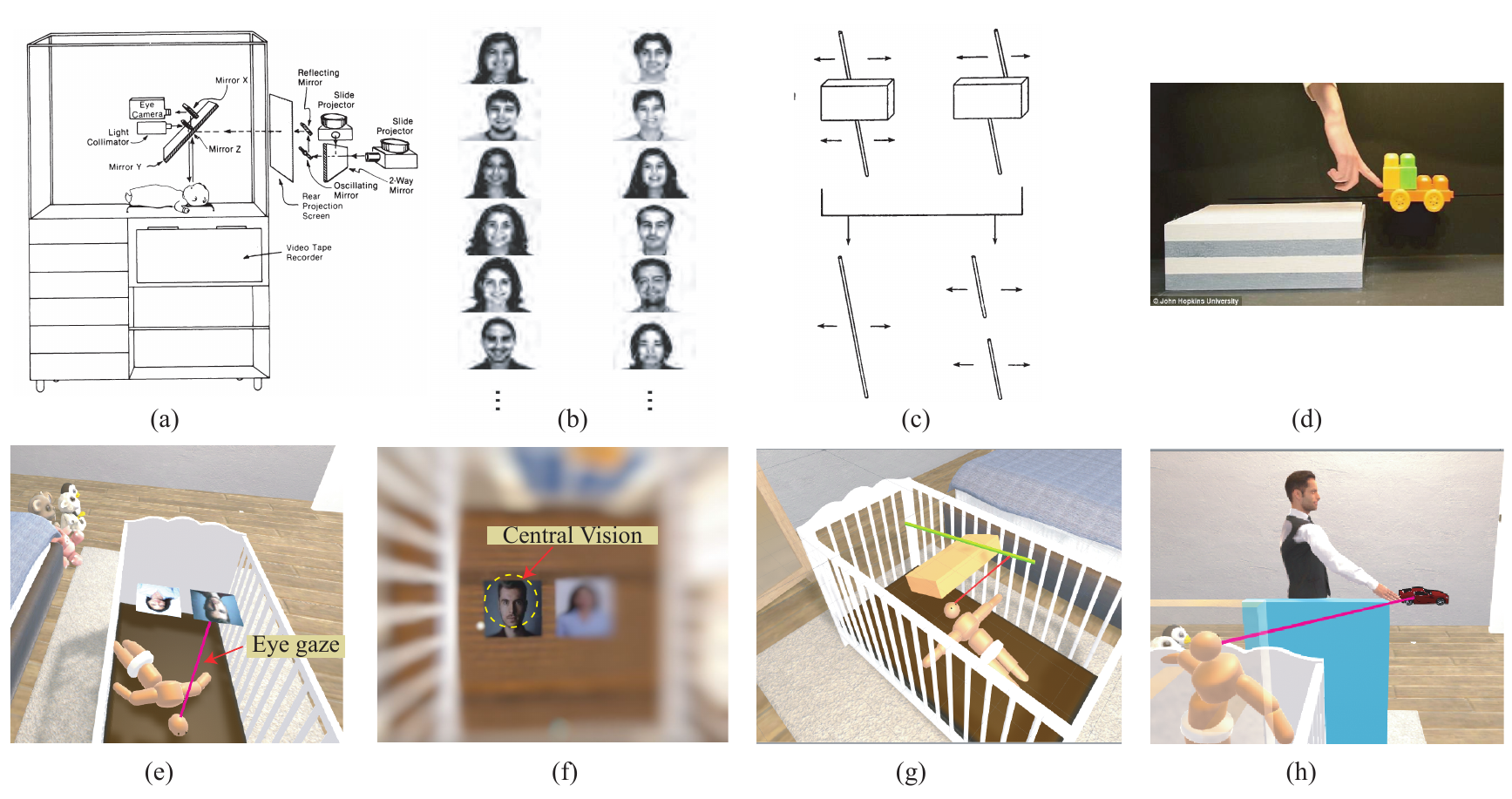}}
  \caption{{\small Example evaluation methods for non-verbal infants and simulated experiments for the artificial agent.  (a) An experiment set up to examine visual pattern according to the stimulus~\cite{haith1988expectation}. A baby views a series of visual stimuli at two or more locations. Visual patterns, such as looking time and looking preference,of the baby are then analyzed. 
  Depending on the developmental stage, they tend to attend more at novel things~\cite{gilmore2002examining}. (b) For example, we can examine if the baby can differentiate between male and female faces~\cite{charlesworth1969role}. (c) When newborn infants, under three months of age, see a rod moving behind a box, they will perceive it as two rods. However, babies, past four months of age, perceive it as a single rod and will be surprised when they are shown two rods~\cite{stahl2015observing}. (e) In our simulated environment, we model eye gaze and central vision. (f) The focused area gives a clear view, while peripheral vision gives a blurred image. (g) Simulated object unity-perception task (h) Test for innate physics. } }
  \label{fig:evaluation}
\end{figure*}

Second generalizable lesson is that we can build upon the prior researches in the developmental psychology to evaluate the developmental progress of non-verbal artificial agent. Because our environment cannot provide sufficient language exposure beyond the first 12 months, the agent cannot acquire advanced language beyond the first few words. 
Consequently, we cannot evaluate the developmental progress of the agent based on their ability to follow verbal instruction or answer questions correctly.  
We overcome this challenge by using studies from developmental psychology. 
There are many experiments revealing developmental milestones for non-verbal infants.
We can simulate and make use of those experiments in SEDRo for developmental assessments.  
As a concrete example, Kellman and Spelke found that babies acquire ~\textit{perceptual completion} around four months using the habituation-dishabituation paradigm~\cite{kellman1983perception}. 
With SEDRo, models can be computationally evaluated by simulating and running experiments to compare behaviors of the agent to the intellectual progress of human infants. 
Fig.~\ref{fig:evaluation} explains these experiments in more detail and shows screenshots of our simulated environment.

The rest of this paper is arranged in the following manner. In section II, we survey related works  which cover different types of simulated environments for developing AI and various evaluation methods for non-verbal agents. Then, in section III, we illustrate our proposed environment, SEDRo. Finally, we draw the conclusion in section IV by pointing out some major limitations of the current version of SEDRo, along with a future plan of actions to resolve these issues.

%% file: sections/background.tex
\section{Background}

We review previous literature for 1) simulated environments for artificial agents  and 2) evaluation methods for non-verbal agents. 

\subsection{Simulated Environments for AI} 
Several environments have been developed for AI research and especially for reinforcement learning researchm~\cite{brockman2016openai, beattie2016deepmind} . The overarching goal was to provide a common benchmark and to lower the barriers of the entry for researchers. Examples include environments in which agents get rewards by following verbal instructions in navigation~\cite{chen2019touchdown, savva2019habitat, chaplot2018gated, hermann2017grounded, shridhar2019alfred}  and give correct answers (question answering)\cite{das2018embodied}.     
Though we've made substantial progress in reinforcement learning with explicit rewards, it is difficult to  transfer these built models  to  develop artificial general intelligence (AGI). 
 Many previous works were conducted to overcome this limitation. 
The difficulty in transferring is mainly because humans do not depend on explicit rewards nor labeled data to learn\cite{hull1943principles, white1959motivation}. 
 Principles such as    intrinsic motivation~\cite{oudeyer2007intrinsic} and free energy~\cite{friston2009predictive} have been proposed  to be the underlying mechanism for learning~\cite{gopnik2004theory, bonawitz2014probabilistic, hawkins2016neurons, nagai2019predictive, schmidt2020self}.  
A number of simulated environments have been proposed to test these hypotheses in the robotics context. 
We can classify previous environments  into artificial environments and human-inspired environments.

\paragraph{Artificial Environments} 
Oudeyer et al. proposed a mathematical formulation for intrinsic motivation and demonstrated similar observations using both simulation and robots~\cite{oudeyer2007intrinsic}. 
Similarly, Haber et al. showed that an agent begins by exploring  an   environment, and then in the later stages begins to interact with objects~\cite{haber2018learning}  based on intrinsic rewards.
These works were conducted on 3D simulated environments built using  game physics engines. 
However, previous works usually focussed on developing and testing a single component of the mechanism, such as self-other cognition, imitation, and joint attention\cite{nagai2019predictive}. 
While it is relatively easy to build artificial environments to test a single component, it is difficult to extend   the environments for multiple tasks. 
To overcome  this limitation, human-inspired environments were also studied.

\paragraph{Human-inspired Environment}
For Human-Inspired Environments, there is a benefit in using infant-like environments. We can be rest assured that those experiences are enough for the development of human-level intelligence. 
Meltzoff et al. elaborated this idea with evidence from developmental psychology, neuroscience, and machine learning~\cite{meltzoff2009foundations}.
The idea of using human-like experience to nurture AI has been actively pursued in the ~\textit{Developmental Robotics (DevRob)} or \textit{Epigenetic Robotics} community~\cite{lungarella2003developmental, asada2009cognitive, cangelosi2015developmental}. 
However, it is challenging to simulate the real world.
Therefore, researchers used 1) physical robots in the real world, or 2)  simulated environment of a simplified real-world, focusing narrow skills. 
Weng et al. developed SAIL robots that explored the world with humans to themselves acquire skills by in navigation and object perception~\cite{weng2001autonomous}.
Later iCub~\cite{metta2008icub}, a  humanoid robot that is modelled after human babies was developed and used for developmental robotics research. Using physical robots, studies on perception and physical behaviors with objects can be conducted~\cite{ruesch2008multimodal,gaudiello2016trust, marocco2010grounding,serhan2019replication}. 
However, physical robots are expensive and providing same experiences for the reproducible research is an open problem. 
To lower the cost of entry for research in robotics, many simulators were developed~\cite{koenig2004design, Webots04}. 
Environment simulations of human development are modelled after stages as early as the fetus~\cite{kuniyoshi2006early, mori2010human} as it is evidenced that fetuses learn auditory~\cite{moon2013language} and sensorimotor coordination.
To tackle the challenge of simulating natural interactions with human users, Murane et al. used virtual reality to allow humans to interact with the robots in the simulation~\cite{murnane2019virtual, murnane2019learning}. 
Using this method, data for human-robot interaction can be accumulated. 

\subsection{Evaluation methods for non-verbal agents}

There are many tests for human-level intelligence, including the Turing test, robot college student test, kitchen tests, and AI preschool test~\cite{adams2012mapping}. 
However, most tests require a capability for language and cannot be used for evaluating progressive intelligence in diverse domains. 
\paragraph{Tests in Developmental psychology} Researchers in developmental psychology developed various evaluation schemes using behavior patterns related to familiarity and novelty.  
These includes visual expectation paradigm~\cite{haith1988expectation}, preferential looking~\cite{fantz1956method}, habituation-dishabituation paradigm~\cite{kaplan1986habituation}, contingent change of the rate in pacifier-sucking behaviors~\cite{moon2013language}. 
For instance, the visual expectations paradigm means that babies look longer and attend more to novel scenes than to familiar scenes. Using these methods, developmental milestones in many skill domains such as visual~\cite{bushnell2001mother}, auditory~\cite{kuhl2007speech}, motor~\cite{clifton1993visually}, social~\cite{maurer1981infants,courage2002infant}, language~\cite{kuhl2007speech}, physics~\cite{kellman1983perception} etc have been studied. 

\paragraph{Psychology-inspired Test for AI} There are previous researches that use human psychological metrics for the evaluation of artificial agents. 
For example, Leibo et al. used human psychology paradigms such as visual search, change detection, and random dot motion discrimination~\cite{leibo2018psychlab}.
However, it tests adult level psychological perception and does not provide developmental milestones. It is also limited to the visual perception domain and does not provide an integrated experience required to learn and perform diverse tasks. 
Piloto et al. suggested the evaluation of physics concepts that are inspired by developmental psychology~\cite{piloto2018probing}. They developed a dataset by examining object persistence, unchangeableness, continuity, solidity, and containment by violation of expectations (VOE) methods. 
The study of complete and diverse tasks at the human level is challenging. 
Crosby et al. used various intellectual animal behaviors in the simulated environment~\cite{crosby2019animal}. 
Tests for ten cognitive categories and a playground that can provide an experience to learn those skills are provided in the work. 
SEDRo builds upon their work to extend those approaches to human-level intelligence.

%% file: sections/plan.tex
\section{Simulating Experience of Human Infants} 

\begin{figure}[t!]
  \centering
  \resizebox{\columnwidth}{!}{\includegraphics{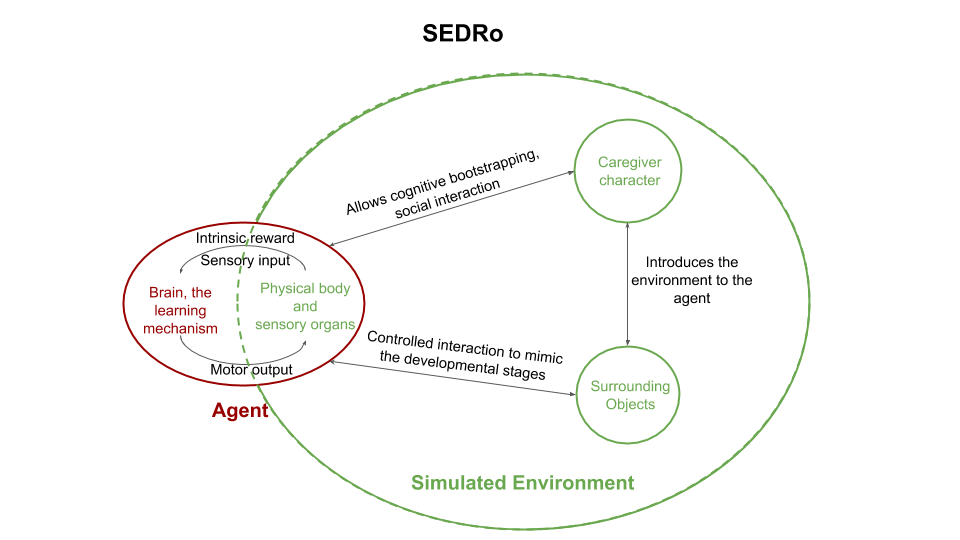}}
  \caption{\small Ecosystem of SEDRo environment.  }
  \label{fig:big_picture}
\end{figure}

In this section, we discuss about the proposed Simulated Environment for Developmental Robotics or SEDRo. Fig.~\ref{fig:big_picture} illustrates the primary components of SEDRo and their inter-relations. 
The two main components in SEDRo are the learning agent (with red border), the simulated environment (with green border). 
Within the simulated environment, there are a caregiver character, surrounding objects in the environment (e.g. toys, cribs, walls etc.) and most importantly the body of the agent. 
The agent will interact with the simulated environment by controlling the muscles in its body according to the sensor signals. 
Interaction between the agent and the caregiver allows cognitive bootstrapping and social-learning, while interactions between the agent and the surrounding objects are increased gradually as the agent gets into more developed stages. 
The caregiver character can also interact with the surrounding objects to introduce them to the agent at the earlier stages of development.

Though there are no rewards that are explicitly awarded by the environment, it does not mean that the reward mechanism does not play a role in the learning. Rather than relying on the environment for the rewards, the responsibility of generating rewards belong to the agent itself. As an example, if an agent can get food from its environment, this input will be given to the agents as a number representing the amount of food in its stomach. It is now the agent's role to generate a negative reward if there is no food in its stomach and positive rewards if new food is given. In this sense, we can say, the body itself is a part of the environment, and what is referred to as the agent is only the brain, which is why the agent’s body in Fig.~\ref{fig:big_picture} is in green font.

\begin{figure}[t!]
  \centering
  \resizebox{\columnwidth}{!}{\includegraphics{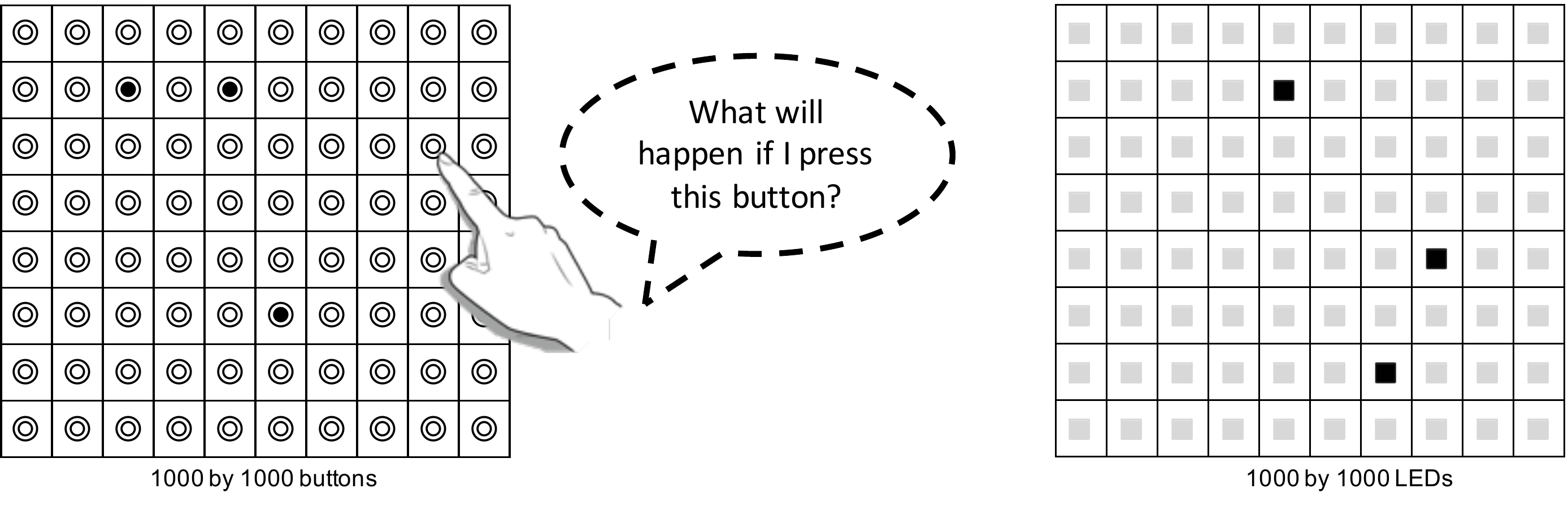}}
  \caption{\small This diagram illustrates our assumption about the learning mechanism. 
  Pressing some buttons in 1,000 by 1,000 button control panel will affect the flickering patterns in 1,000 by 1,000 LEDs panel. 
 The button panel represents the motor output vector, and the LED panel represents the sensory input vector. 
  Please note that there are no labels on those two vectors and the learning mechanism needs to learn how to operate the body.  }
  \label{fig:model}
\end{figure}

SEDRo provides the diverse experiences of human infants from the stage of a the fetus to first 12 months of life. 
A new-born brain must learn to control its body. We can compare  this with trying to learn to operate a machine using a control panel of 1,000 by 1,000 LEDs and  1,000 by 1,000 buttons.  To make this even more challenging, LEDs and buttons are not labeled  as shown in Fig.~\ref{fig:model}. Each LED blinks, maybe sparsely. If you push some buttons, the blinking pattern of the LEDs seems to change and sometimes not; is not easy to track. You need to make sense out of this huge matrix of buttons and LEDs that Piaget called sensorimotor stage~\cite{piaget1952origins}. The role of a brain model is to compose an output behavior vector given a sensor vector.

\subsection{Curriculum for Development}

To make the learning easier, human infants develop in a curriculum which scaffolds the involved sensory and motor capabilities.~\cite{smith2018developing, turkewitz1982limitations, berlyne1960conflict,mirvis1991flow}. For example, in the fetus stage there are no visual inputs. A small subset of LEDs and buttons that are available at that stage can be isolated to master new skills such as sucking a thumb or rotating body. In the first three months, babies are very near-sighted and do not have any mobility, which makes many visual signals stationary. At later stages, when babies learn to sit and grasp, they develop alternative strategies of learning using a rotating viewpoints and the contingent verbal speech of caregivers. 

In SEDRo, the input output signal changes according to the development of the agent. For example, the agent in the womb stage will not have any visual input signals which will be available after birth. But for the first 3 months, visual signals will represent nearsightedness. Muscles will develop over time. Full force at the early stage will not be enough for an agent to crawl or stand, but it will steadily increase to afford walking in the later stages.  

\begin{figure}[b!]
  \centering
  \resizebox{0.9\columnwidth}{!}{\includegraphics{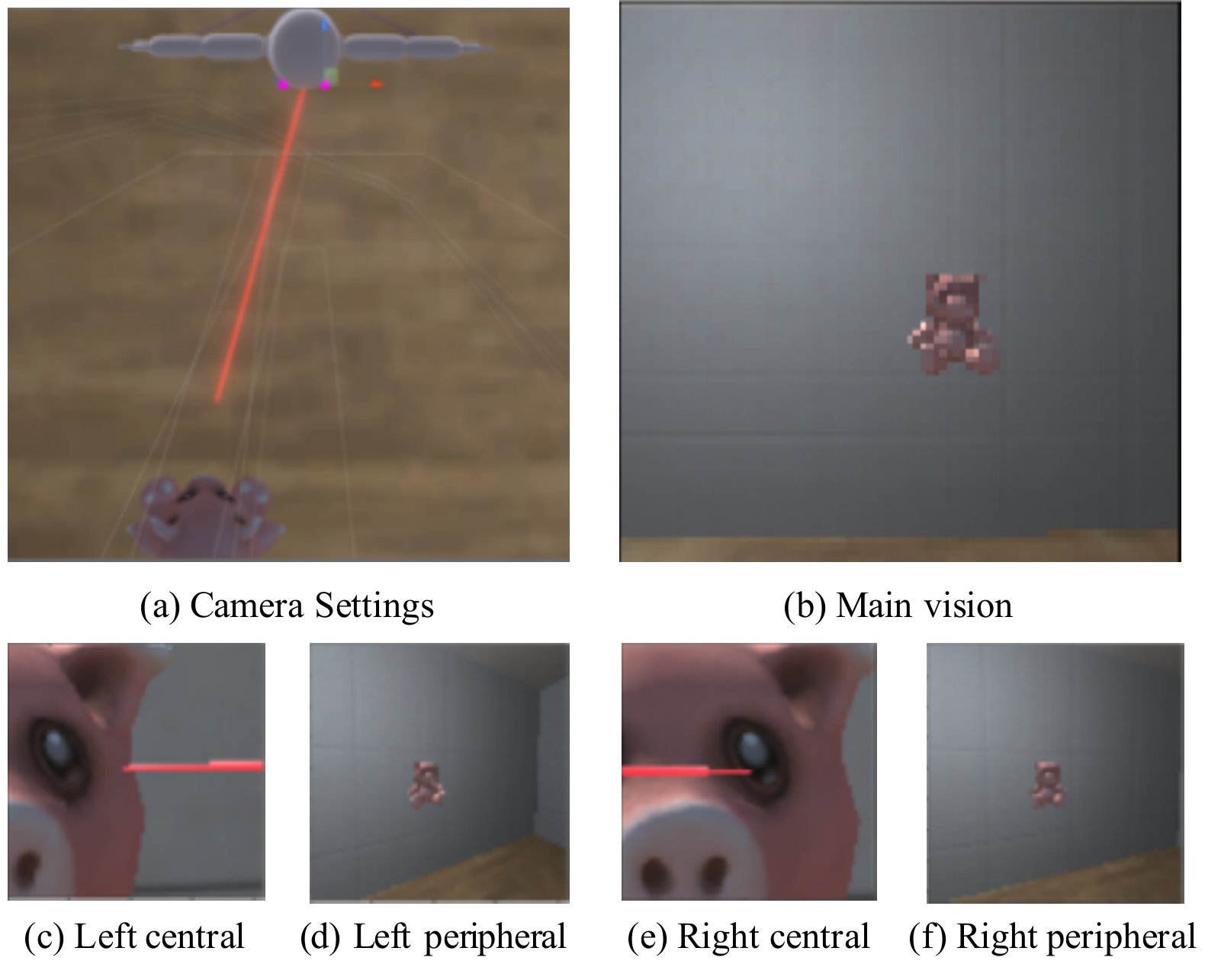}}
  \caption{\small The visual system of the agents in SEDRo. The orange laser beam in (a) shows visual attention. Each eye has a central (c and e) and a peripheral vision (d and f).   For debugging purpose, a main view is provided as shown in (b). }
  \label{fig:vision}
\end{figure}

\subsection{Specification for I/O Vectors}
The sensory input consists of touch, vision, acceleration, gravity, and proprioceptors.  
Visual attention plays an important role in the evaluation of non-verbal human infants as explained in Sec.~\ref{sec:evaluation}. In SEDRo, the artificial infant can control eye movement with three parameters - vertical angle, horizontal angle, and focal length. 
To simulate central and peripheral vision with two eyes, four images are generated as central and peripheral vision images for left and right eyes each. 
Central vision has an 8\degree ~field of view (FOV) and a higher resolution, while the peripheral vision has a 100\degree ~FOV ~and a lower resolution. One additional image is provided for the purpose of debugging that will represent reconstructed visual imagery in the brain.   
Five cameras with different settings have been used in the game engine for implementation. 

The touch sensors are spread all over the body but the distribution pattern varies. The face, lips, and hands have a higher density of touch sensors than the torso. In the current version, there are 2,110 touch sensors. Each body part is  segmented into meshes and a touch sensor provides a binary feature which represents if a contact has been made. 
We have implemented touch feature using collision information provided by the game engine. 

The motor output vectors constitute muscle torques, which will determine the 53 motors, including the 9 degree of freedom (DOF) in each hand inspired by iCub~\cite{metta2008icub}. 
Main loop of the environment runs 100 steps per second motivated by the human biological brain~\cite{hawkins2004on}. At each step, the agent will read a sensory input vector and write a motor output vector.


\subsection{Social Event Scenario}
Social interaction plays an important role for human development. In SEDRo, we are building scenarios for social interaction according to the following process:
\begin{enumerate}
  \item We start by choosing a meaningful interaction pattern by reviewing developmental psychology literature.
  \item We write a scenario for the chosen interaction. 
  \item Actors perform that scenario and and we capture their behaviors using a motion capture facility. 
  \item We add the recorded scenario into SEDRo along with a schedule for the event. 
\end{enumerate}

Building a library of the social event is time consuming and we anticipate the SEDRo environment will expand over the coming years. We will maintain the versioning of SEDRo such that the research using SEDRo can be reproducible.  

\begin{table*}   
  \small
  \caption{Summary of Developmental Milestones (M represents months after birth)}
\begin{center}
  \begin{tabular}{|m{1.1cm} || m{3.4cm}| m{3.4cm} | m{3.4cm} | m{3.4cm}|} 
 \hline
\thead{ Stage} & \thead{Fetus Stage} & \thead{Immobile Stage \\ (Less than 3 Months)} & \thead{Crawling Stage \\ (4-10 Months)} & \thead{Walking Stage\\ (11-18 Months)} \\ [0.5ex] 
 \hline\hline
 \makecell{Descri\\ption}
 & No vision

 & Near sighted vision.  
 & Fully developed vision. Sit and interact with objects. Interact with other persons by babbling.
& Fully developed muscles. First words
 \\ 
 \hline
 Vision & 

 & Visual expectation(0 vs 3M)~\cite{haith1988expectation, canfield1991young,adler2008infants,wentworth2002spatiotemporal}, face preference(1 vs 2M)~\cite{maurer1976developmental,haan2002specialization,maurer1981infants, morton1991conspec}, face preference(2~3 days)~\cite{bushnell2001mother, bushneil1989neonatal}
 , gender detection(0 vs 3M)~\cite{leinbach1993categorical, quinn2002representation}, depth perception (0 vs 2M)~\cite{campos1970cardiac}

 & Visual scan pattern (2 vs 11 weeks) ~\cite{bronson1991infant}, tracking occluded objects(4 vs 6 Months)~\cite{johnson2003development}, lost ability to distinguish faces of different gender(3 vs 9 M)~\cite{quinn2002representation}    
 
 & Novelty preference inversion (6-12 months)~\cite{roder2000infants, colombo2006emergence} \\
 \hline
\makecell{Joint\\atten\\tion}
 
 & 
 
 & left/right attention manipulation
 & Gaze angle detection, fixation of first salient object
 
 & Mutual gaze through eye contact ~\cite{kaplan2006challenges}, Fixation of any salient object, declarative pointing, drawing attention \\
 
 \hline
 Motor 
 
 & Hand/face contacts (11 gestation weeks)~\cite{mori2010human} 
 
 & Open hand grasping~\cite{von1984developmental,white1964observations}
 
 & Recognizing own motion vs others(3 vs 5 Months)~\cite{bahrick1985detection,gergely1999early}
 
 & Partial integration of visual and motor skill (9 Months)~\cite{berthier2001using} \\

\hline
 \makecell{Lang\\guage} & & Differentiate mother tongue and foreign language~\cite{moon2013language}, marginal babbling & canonical babbling & intentional gestures, single words, word-gesture combination~\cite{hoff2013language}\\
 \hline
 \makecell{Reason\\ing} 
 
 &  
 
 & Self-perception at mirror(3 Months)~\cite{courage2002infant}
 
 & Fear of heights (after crawling)\cite{campos1992early,kermoian1988locomotor}, Allocentric spatial frame of reference (9 Months)~\cite{acredolo1978development,acredolo1980developmental,acredolo1984role}
 
& Mark test(15 Months)~\cite{lewis2012social}, adapted use of hook(12 Months)~\cite{van1994affordances}\\

 \hline
\end{tabular}

\end{center}
\end{table*}

\subsection {Evaluation Framework for Non-verbal Agents}
\label{sec:evaluation}
We have developed an evaluation framework for the development of skills in multiple domains by simulating established experiments from developmental psychology.   
There are multiple developmental milestones in multiple skill domains.
At each stage, there are key milestones that the agent needs to satisfy. 
Researchers may choose to replay relevant experiences if the agent does not achieve those milestones.
Consequently, the agent will experience an adaptive experience based on its current capability rather than experiences based on a fixed time schedule. 
Fig.~\ref{fig:evaluation} shows example tasks in the developmental psychology and screenshots of preliminary prototypes simulating those experiments. 

In current version, we developed a visual expectation paradigm experiment with a moving rod. 
The visual attention pattern over the moving rod can be acquired as a separate channel in the gym interface. 
Table 1 summarizes our plan for evaluation experiments in domains such as vision, motor, attention, and reasoning. 
Each evaluation has a different expected behavior pattern between two stages of human development. For example, two month old infants cannot predict regular pattern, but at 3.5 months, infants exhibit anticipatory eye movement 200 ms before the actual pattern visual expectation~\cite{haith1988expectation, canfield1991young,adler2008infants,wentworth2002spatiotemporal}. 
We leverage such known developmental milestones to develop suites of simulated experiments for evaluating the development of the artificial agent.
The evaluation will conduct multiple experiments and compare the results with those of the human participants.

\subsection{Implementation Detail}

We use Unity3D 2018.4 for the development of the environment. Unity ML agent ~\cite{juliani2018unity} is used to implement Open AI gym interface~\cite{brockman2016openai}. To record behaviors of the actors, we use Motive Body software with Opti-track motion capture system with ten Prime 17W cameras.

%% file: sections/discussion.tex
\section{Discussion}

As SEDRo is a work in progress, here we discuss its limitations, a few alternative approaches and future works.

\subsection{Limitations}

A major limitation of our work is the lack of back and forth interactive conversation between the caregivers and the infant agent. 
Currently, only two types of conversations are supported;

\begin{itemize}
    \item Caregiver initiated conversation that will be played according to a pre-determined schedule, and 
    \item Contingent response that will be played conditioned on infant agent behaviors such as cooing or touching toys. 
\end{itemize}

Despite building diverse scenarios for these two conversation types is a challenge by itself due to the sheer number of the required diverse experience, they would not provide enough experience for the development of language acquisition beyond the first year level. 
One potential approach to overcome this limitation would be to ask humans to interact with the artificial agent using a virtual reality technique~\cite{murnane2019learning,murnane2019virtual}. 
Another option would be to use a physical embodied robot and ask humans to take care of it. 
We claim that SEDRo can be used to test cognitive architectures before the need to perform physical robot experiments, thereby helping in reducing the number of candidate architectures for the expensive physical robot experiments.

In SEDRo, we simulate the human infant experiences, but an  alternative is to use a completely artificial environment that is not relevant to human experience but still requires skills in many domains. 
For example, emergent communication behaviors were observed in the reinforcement learning environment with multiple agents~\cite{eccles2019biases,cao2018emergent, das2018tarmac,foerster2016learning}. 
Through similar researches, though we might find clues about the underlying human learning mechanism, it might be challenging to apply them to human robot interaction because language is a set of arbitrary symbols shared between members~\cite{kottur2017natural}.

Another possibility is to transform existing resources into an open-ended learning environment. 
Using Youtube videos to create a diverse experience is an example. 
However, Smith and Slone pointed out that these kinds of approaches use shallow information about a lot of things, whereas, on the contrary, human infants begin by learning a lot about a few things~\cite{smith2017developmental}. 
In addition to that, visual information from the first years of human life constitutes an egocentric view of the world. The allocentric view emerges only later, after 12 months of age. 
Furthermore, humans rely heavily on social interactions to learn.
While infants can learn a language by being tutored by an instructor, they cannot learn by seeing a recorded video of an same tutoring~\cite{kuhl2007speech}.  
Therefore we think that certain necessary skills have to be acquired before learning from those resources becomes feasible.

\subsection{Conclusion and Future Work}
We are building SEDRo, an environment that simulates the early experiences of a human from the stage of a fetus to 12 months of age. 
The open-ended and unsupervised nature of the environment requires agents to avoid fitting to specific tasks. 
To evaluate the development of intelligent behaviors of non-verbal artificial agents, a set of experiments in developmental psychology will be simulated in SEDRo.
We expect researchers in the AI and robotics community to discover the learning mechanism for artificial general intelligence by testing different cognitive architectures using the open-ended learning environment developed in our project. 